%% file: main.tex
\documentclass[10pt,twocolumn,letterpaper]{article}

\usepackage{cvpr}              
\usepackage{marvosym}
\usepackage{chngcntr}
\usepackage[accsupp]{axessibility}  

\newcommand{\revision}[1]{#1}

\input{preamble}

\definecolor{cvprblue}{rgb}{0.21,0.49,0.74}
\usepackage[pagebackref,breaklinks,colorlinks,allcolors=cvprblue]{hyperref}

\def\paperID{19460} 
\def\confName{CVPR}
\def\confYear{2026}

\title{From Exploration to Exploitation: A Two-Stage Entropy RLVR Approach for Noise-Tolerant MLLM Training}

\author{
    \textbf{Donglai Xu}$^{1}$\thanks{Equal contribution.},
    \textbf{Hongzheng Yang}$^{2}$\footnotemark[1],
    \textbf{Yuzhi Zhao}$^{3}$\thanks{Corresponding author.},
    \textbf{Pingping Zhang}$^{3}$,
    \textbf{Jinpeng Chen}$^{3}$,\\
    \textbf{Wenao Ma}$^{2}$, 
    \textbf{Zhijian Hou}$^{3}$, 
    \textbf{Mengyang Wu}$^{2}$, 
    \textbf{Xiaolei Li}$^{4}$,
    \textbf{Senkang Hu}$^{3}$,\\
    \textbf{Ziyi Guan}$^{5}$,
    \textbf{Jason Chun Lok Li}$^{5}$,
    \textbf{Lai-Man Po}$^{3}$
    \\ 
    \fontsize{11pt}{12pt}\selectfont
    $^1$Independent Researcher~~~
    $^2$The Chinese University of Hong Kong~~~ 
    \fontsize{11pt}{12pt}\selectfont
    $^3$City University of Hong Kong~~~\\
    $^4$Hong Kong University of Science and Technology~~~
    $^5$University of Hong Kong
     \\ 
     \fontsize{11pt}{12pt}\selectfont
    donglaixu99@gmail.com; hzyang22@cse.cuhk.edu.hk; yzzhao2-c@my.cityu.edu.hk
}

\begin{document}
\maketitle

\input{sec/0_abstract} 

\begin{figure*}[t]
\centering
\includegraphics[width=1\linewidth]{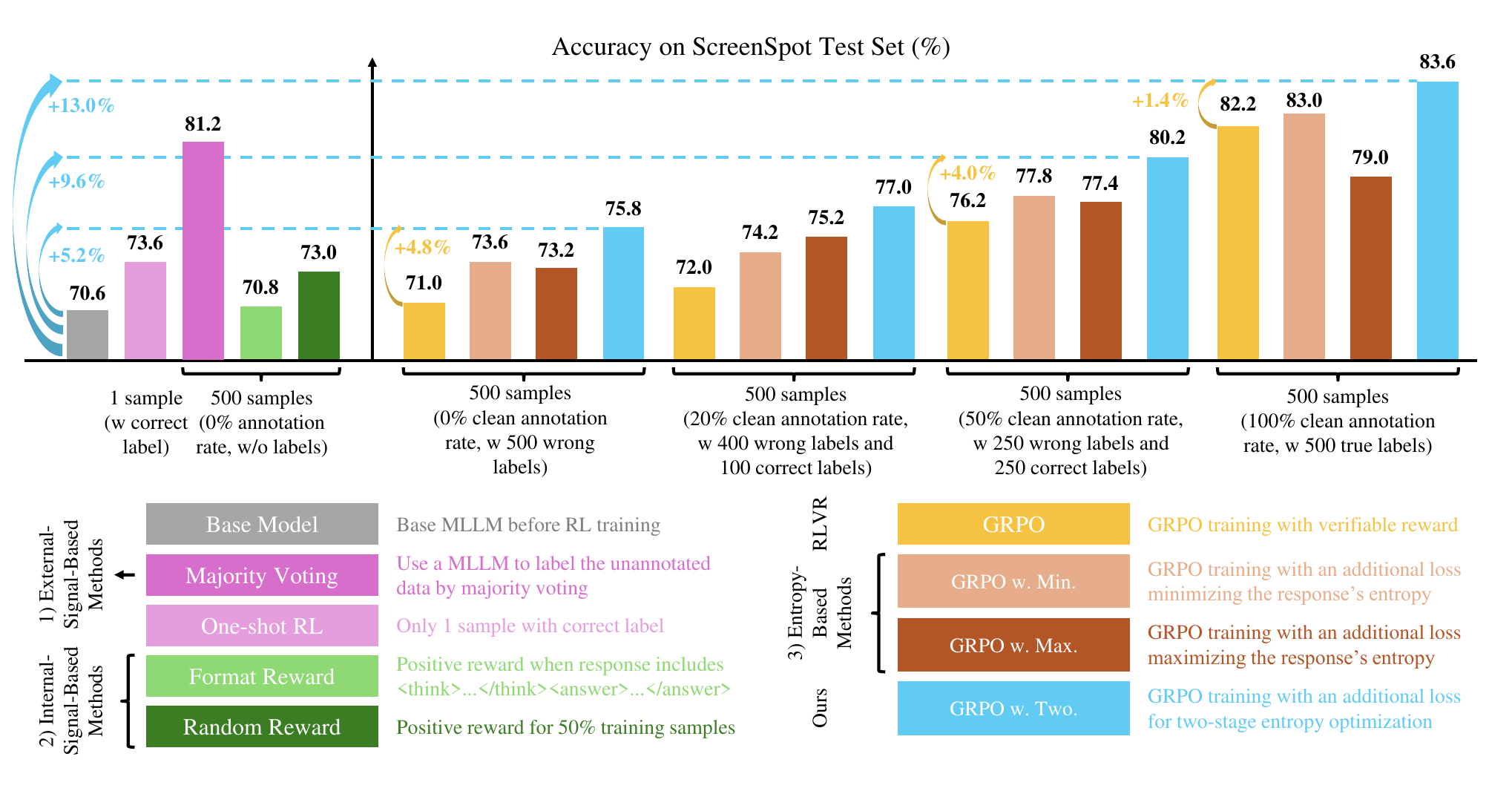}
\centering
\caption{ScreenSpot accuracy after 1000 steps of different training strategies on Qwen2.5-VL-3B model. The horizontal axis includes different training data configurations. The proposed two-stage entropy-guided RLVR training method (GRPO w. Two.) performs better than one-shot RL \cite{gao2025oneshotentropyminimization}, RLVR with “spurious rewards” (including format reward and random reward) \cite{shao2025spuriousrewardsrethinkingtraining}, and RLVR with pure entropy minimization or maximization \cite{zhang2025right} \revision{on diverse noisy label settings, with clean annotation rates ranging from (0\%, 20\%, 50\%, 100\%).}}
\label{cfa}
\end{figure*}

\input{sec/1_intro}
\input{sec/2_related}

\input{sec/3_preliminary}

\input{sec/4_method}

\input{sec/5_experiment}

\input{sec/6_discussion}

{
    \small
    \bibliographystyle{ieeenat_fullname}
    \bibliography{main}
}

\appendix
\input{sec/X_suppl}

\end{document}

%% file: preamble.tex
\usepackage{times}  
\usepackage{helvet}  
\usepackage{courier}  
\usepackage[hyphens]{url}  
\usepackage{graphicx} 
\usepackage{natbib} 
\usepackage{caption} 
\usepackage{algorithm}

\usepackage{algpseudocode}
\usepackage{amsmath}
\usepackage{amssymb}

\usepackage[utf8]{inputenc}
\usepackage{multirow}

\usepackage{newfloat}
\usepackage{listings}

\usepackage{booktabs}
\usepackage{amsmath}
\usepackage{amsfonts}

%% file: sec/0_abstract.tex
\begin{abstract}
Reinforcement Learning with Verifiable Rewards (RLVR) for Multimodal Large Language Models (MLLMs) is highly dependent on high-quality labeled data, which is often scarce and prone to substantial annotation noise in real-world scenarios. 
\revision{Existing RLVR methods under noisy supervision can overfit to incorrect labels and suppress the response diversity essential for the reward ranking signal in Group Relative Policy Optimization (GRPO).}
To address these challenges and enhance noise tolerance, we propose a two-stage token-level  entropy optimization method for RLVR. This approach dynamically guides the model from exploration to exploitation during training. In the initial exploration phase, token-level entropy maximization promotes diverse outputs, serving as a regularizer that \revision{mitigates} premature convergence to noisy labels and \revision{ensures} sufficient intra-group variation, enabling more reliable \revision{advantage} estimation in GRPO. As training progresses, the method transitions into the exploitation phase, where token-level entropy minimization encourages the model to produce confident outputs, thereby consolidating acquired knowledge and refining prediction accuracy. Empirically, across \revision{diverse} MLLM backbones, various noise settings, and multiple tasks, \revision{our phased entropy schedule delivers superior overall robustness and outperforms representative external-signal, internal-signal, and entropy-based baselines.}
\end{abstract}

%% file: sec/1_intro.tex
\section{Introduction}
\label{sec:intro}

Recently, Reinforcement Learning with Verifiable Rewards (RLVR) has gained recognition for its effectiveness, as evidenced by its superior generalization compared to supervised fine-tuning (SFT)~\cite{chusft_verse_rl}, its ability to elicit reasoning, and its ease of implementation. A notable example is Group Relative Policy Optimization (GRPO) \citep{guo2025deepseek}, applied by DeepSeek-R1 \citep{guo2025deepseek}, which exemplifies these strengths. RLVR has demonstrated success across a wide range of domains, including mathematical reasoning \citep{yang2024qwen2, shao2024deepseekmath, nguyen2026adaptive}, formal verification \citep{xin2024Deepseek-prover-v1.5, ren2025Deepseek-prover-v2}, and code generation \citep{wei2025swe}. Moreover, RLVR has been extended to multimodal tasks, enhancing the reasoning capabilities of Multimodal Large Language Models (MLLMs). These applications span image classification and object grounding \citep{liu2025visualrftvisualreinforcementfinetuning, huang2025visionr1, shen2025vlm, chen2025r1v}, image segmentation \citep{liu2025seg}, medical reasoning \citep{lai2025medr1}, video understanding \citep{wang-2025-open-r1-video-web, feng2025video}, and graphical user interface (GUI) reasoning \citep{lu2025uir1enhancingefficientaction, luo2025guir1generalistr1style}. Despite these advancements, a critical challenge remains: RLVR methods typically rely on high-quality labeled data to compute verifiable rewards. In real-world scenarios, datasets are frequently accompanied by annotation noise, posing a significant barrier to effective RLVR implementation.

To address the challenge of applying RLVR to datasets with annotation noise, recent methodologies can be grouped into three primary categories:

1. \textbf{External-Signal-Based Methods}: These approaches utilize external verifiable signals to guide RLVR training, such as compilers for code generation \citep{li2025autotriton}, Large Language Models (LLMs) as evaluators (e.g., LLM-as-a-Judge) \citep{gu2024survey_llmasajudge}, and Test-time Reinforcement Learning (TTRL) \citep{wei2025unsupervised, zuo2025ttrl}. These methods exhibit inconsistent performance due to variations in LLM capabilities across domains, and tools like compilers are often task-specific, limiting their applicability.

2. \textbf{Internal-Signal-Based Methods}: These methods derive rewards directly from model outputs, such as random rewards or format rewards. \revision{These internal reward signals do not rely on labeled data or external tools \citep{shao2025spuriousrewardsrethinkingtraining}.} Although these approaches offer flexibility, their effectiveness is constrained, as the \revision{internal} reward \revision{signals} are often not closely aligned with task-specific objectives.

3. \textbf{Entropy-Based Methods}: These methods leverage entropy to guide training. For example, \citet{wang2025reinforcementlearningreasoninglarge} proposed a one-shot RL scheme that achieves significant improvements in mathematical reasoning using the entropy loss. Similarly, \citet{zhao2025learningreasonexternalrewards} employed self-certainty signals, while EMPO \citep{zhang2025right} minimized predictive entropy directly. However, these approaches often overemphasize entropy reduction, potentially overlooking the dynamic role of entropy across different training stages.

To investigate the robustness of RLVR under noisy data conditions, we evaluate the performance of MLLMs trained with different RL methods on two visual tasks: GUI grounding and fine-grained classification. We systematically vary the proportion of mislabeled data while maintaining a fixed training set size. The results for the GUI grounding task are presented in Figure~\ref{cfa}. Our key observations regarding the three methodological categories are as follows:

1. As the proportion of mislabeled data decreases, \revision{the} model accuracy \revision{after training} generally increases.
\revision{External-signal-based methods, such as TTRL~\citep{zuo2025ttrl}, rely heavily on MLLMs for pre-labeling.
The capability of the MLLM used for pre-labeling directly affects the ratio of noisy data introduced into the training set, which consequently imposes an upper bound on the final model performance.}



2. With a small proportion of correctly labeled data, standard GRPO training outperforms internal-signal-based methods, such as those relying on spurious rewards \citep{shao2025spuriousrewardsrethinkingtraining}.

3. Augmenting GRPO with entropy-based losses \citep{zhang2025right} consistently yields superior performance compared to using GRPO alone. Similar trends are observed across other vision tasks.



\revision{Based on these observations, we find that standard GRPO alone can already match or exceed the performance of internal-signal-based methods under moderate noise conditions (i.e., excluding purely random noise). 
Furthermore, augmenting standard GRPO with entropy-based methods can yield additional improvements against noisy annotations. However, if the optimization objective is naively reduced to either only entropy maximization or only entropy minimization}, the learning dynamics can become problematic.
\revision{Pure entropy maximization makes convergence difficult}, while exclusive entropy minimization may trap the model in sub-optimal deterministic behaviors, especially facing the label noise. Furthermore, pure entropy minimization suppresses the response diversity, which is necessary for the informative advantage estimation required by GRPO. 
We argue that entropy optimization should be scheduled and switched between the two regimes, which could offer a controlled trade-off between exploration and exploitation without sacrificing convergence stability.

Specifically, we propose a two-stage entropy-guided RLVR training method. During the early phase of training, \revision{we maximize token-level entropy to encourage more diverse outputs.} This promotes exploration and \revision{mitigates} overfitting to noisy labels. \revision{As training progresses}, the model has captured most of the information from the datasets. We then proceed to the second stage, where entropy minimization is applied to encourage more confident and deterministic output. By explicitly guiding the model from exploration to exploitation, this two-stage method enhances the model's ability to learn from noisy datasets. For instance, by applying the two-stage entropy optimization to Qwen2.5-VL-3B \cite{bai2025qwen25vltechnicalreport} with 50\% noisy labels, the method further boosts performance from 76.2\% to 80.2\% on ScreenSpot dataset \cite{cheng2024seeclickharnessingguigrounding}, with similar gains observed across other levels of label noise (e.g., from 71\% to 75.8\% for 100\% noisy labels, and from 82.2\% to 83.6\% for 0\% noisy labels), as shown in Figure 1.
It also outperforms pure entropy maximization or minimization in most noise settings, yielding the strongest overall robustness.
Our contributions can be summarized as follows:
\begin{itemize}
    \item We conduct comprehensive experiments across multiple dimensions: 1) varying \revision{noisy} annotation rates, 2) diverse model architectures and scales (Qwen2-VL-2B, Qwen2.5-VL-3B, Qwen2-VL-7B~\citep{wang2024qwen2vl}, \revision{InternVL-3.5-2B~\citep{wang2025internvl3}}), and 3) multiple tasks (GUI grounding, fine-grained classification, and open-vocabulary object detection), to evaluate the impact of noisy labels on RLVR.

    \item \revision{We identify the limitations of existing RLVR methods under noisy supervision and introduce a two-stage entropy-guided optimization method. By transitioning from exploration to exploitation, our approach mitigates overfitting to noisy labels while consolidating knowledge.}

    \item \revision{Our phased entropy strategy outperforms standard GRPO and existing entropy-based methods across different models, task types and noise conditions.}
\end{itemize}


%% file: sec/2_related.tex
\section{Related Works}

\subsection{Reinforcement Learning with Verifiable Rewards}

RLVR leverages verifiable signals to compute rewards, particularly for tasks with well-defined correctness criteria, such as mathematical reasoning and code generation \citep{shao2024deepseekmath, lambert2024tulu, hu2025open, team2025kimi}. Unlike traditional reinforcement learning approaches that rely on learned reward models, RLVR employs rule-based verification functions, such as exact answer matching, to mitigate the complexities and potential biases associated with learned rewards. This characteristic has enabled RLVR to achieve state-of-the-art reasoning capabilities in LLMs, as exemplified by DeepSeek-R1 \citep{guo2025deepseek}. The GRPO algorithm and its variants \citep{shao2024deepseekmath} have further extended RLVR to multimodal scenarios, including image classification \citep{liu2025visualrftvisualreinforcementfinetuning}, geometry reasoning \citep{huang2025visionr1}, GUI grounding \citep{luo2025guir1generalistr1style}, and multi-step reasoning tasks such as search \citep{jin2025search}. Despite these successes, RLVR's effectiveness is limited to domains with reliable verifiable signals and high-quality annotations, posing challenges in scenarios with noisy data.

\subsection{Reinforcement Learning with \revision{Noisy} Annotations}


\revision{In scenarios where accurate and clean data is unavailable, existing methods often have to utilize noisy annotations to guide RLVR training.} LLM-as-a-Judge~\citep{yuan2024self,xiong2025self} is a well-known method, which utilizes the LLM itself as a \revision{noisy reward signal} when \revision{accurate} human feedback is not available. Recently, TTRL~\citep{zuo2025ttrl} employs majority voting across diverse model outputs to generate \revision{noisy pseudo labels}, which serve as verifiable rewards to enhance mathematical reasoning through RL training.
Additionally, research on spurious rewards \citep{shao2025spuriousrewardsrethinkingtraining} has explored format rewards and random rewards.  These internal reward signals do
not rely on any labeled data, either with clean or noisy annotations.
The majority of these studies have focused on math reasoning and code generation tasks with either pure unlabeled data or partially labeled data with clean annotations. 
In this work, we systematically evaluate the impact of these reward signals on multimodal tasks under noisy supervision.

\subsection{Entropy in Reinforcement Learning}



Recently, entropy minimization~\citep{grandvalet2004semi} has been adapted to RLVR~\citep{zhao2025learningreasonexternalrewards, wang2025reinforcementlearningreasoninglarge}. For instance, \citet{zhao2025learningreasonexternalrewards} only utilized self-certainty as a reward signal in RL training, achieving superior out-of-domain performance and matching standard GRPO training on mathematical reasoning benchmarks. Similarly, the EMPO framework \citep{zhang2025right} minimizes the entropy of output sequences, leveraging internal model consistency as an effective reward signal. Additionally, Seed-GRPO \citep{chen2025seed} employs entropy to modulate the magnitude of policy updates, enhancing training stability. 


\revision{However, existing approaches primarily utilize pure entropy as a standalone reward signal, or focus on improving training stability under partially labeled datasets with strictly clean annotations. In contrast, our work investigates the dynamic role of entropy-based mechanisms for multimodal tasks under noisy supervision. }

%% file: sec/3_preliminary.tex
\section{Preliminary}

\subsection{Group \revision{Relative} Policy Optimization (GRPO)}

RLVR leverages binary rewards for policy optimization. Unlike traditional reinforcement learning approaches that rely on human feedback or learned preference models, RLVR employs rule-based verification labels, such as exact answer matching, compiler feedback, or mathematical correctness checks to determine reward assignment.

GRPO serves as the primary algorithm for RLVR training. The GRPO training process begins by sampling $K$ responses $\{y_1, y_2, \ldots, y_K\}$ from the current policy $\pi_\theta(\cdot|x)$ for each input $x$. Each response $y_i$ is evaluated using a verifiable reward function $\mathcal{R}(y_i, y^*)$ that returns a binary signal based on correctness verification. The key innovation of GRPO is its group-wise advantage estimation that normalizes rewards within each group to reduce variance.
For a given group of $K$ responses with rewards $\{r_1, r_2, \ldots, r_K\}$, GRPO computes the advantage for each response as:

\begin{equation}
A_i = \frac{\revision{r_i} - \text{mean}(r(y_{1:K}))}{\text{std}(r(y_{1:K}))},
\end{equation}
where $\text{mean}(r(y_{1:K}))$ and $\text{std}(r(y_{1:K}))$ are the mean and standard deviation of rewards within the group, respectively.

The policy gradient becomes:
\begin{equation}
\nabla_\theta \mathcal{L}_{\text{GRPO}} = -\mathbb{E}_{x \sim \mathcal{D}} \left[ \sum_{i=1}^{K} \sum_{t=1}^{T_i} A_i \nabla_\theta \log \pi_\theta(y_{i,t}|y_{i,<t}, x) \right].
\end{equation}
In practice, we use a clipped surrogate objective to constrain the update relative to the old policy and avoid overly aggressive parameter changes.
\begin{equation}
\begin{aligned}
\mathcal{L}_{\text{GRPO}}(\theta) = -\mathbb{E}_{x \sim \mathcal{D}} \Bigg[ \sum_{i=1}^{K} \sum_{t=1}^{T_i} 
\min\Bigg(
    \frac{\pi_\theta(y_{i,t}|y_{i,<t},x)}{\pi_{\theta_{\text{old}}}(y_{i,t}|y_{i,<t},x)} A_i, & \\
    \text{clip}\left( 
    \frac{\pi_\theta(y_{i,t}|y_{i,<t},x)}{\pi_{\theta_{\text{old}}}(y_{i,t}|y_{i,<t},x)}, 
    1-\epsilon, 1+\epsilon \right) A_i 
\Bigg)
\Bigg],
\end{aligned}
\label{grpo_loss}
\end{equation}
where $T_i=|y_i|$ is the response length.

%% file: sec/4_method.tex
\section{Methodology}

\subsection{Token-Level Entropy}
The foundation of our approach lies in leveraging token-level entropy as a granular measure of uncertainty in text generation. Unlike sequence-level entropy, which captures the overall uncertainty of an output, token-level entropy quantifies the predictability of each token at every generation step. Formally, for an input sequence $x$ and partially generated tokens $y_{<t}$, the model produces a conditional probability distribution $\pi_\theta(v \mid x, y_{<t})$ over vocabulary $V$. The per-token entropy is computed as:
\begin{equation}
\mathcal{H}_t(x, y) = -\sum_{v \in \mathcal{V}} \pi_\theta(v \mid x, y_{<t}) \log \pi_\theta(v \mid x, y_{<t}).
\end{equation}

The token-level entropy for the entire sequence is then computed by averaging over all $T$ tokens in the trajectory:
\begin{equation}
\mathcal{H}_{\text{token}}(x,y) = \frac{1}{T} \sum_{t=1}^{T} \mathcal{H}_t(x,y).
\end{equation}
where $T=|y|$ is the response length.
The corresponding entropy loss is then defined as:
\begin{equation}
\mathcal{L}_{\text{entropy}} = - \mathbb{E}_{x \sim \mathcal{D}} \left[ \frac{1}{K} \sum_{i=1}^{K} \mathcal{H}_{\text{token}}(x, y_i) \right].
\label{entropy_loss}
\end{equation}
where $K$ is the number of responses sampled per input $x$.

\subsection{Two-Stage Entropy-Guided GRPO}

\revision{The role of entropy in learning has been studied from the complementary perspectives of exploration and exploitation.}
Early work in semi-supervised classification~\cite{lee2013pseudo_label, grandvalet2004semi} argues that optimizing the predictive distribution towards low entropy transforms unlabeled inputs into effective constraints on the classification decision boundary.  
Deep reinforcement learning~\cite{haarnoja2018soft_entropy_max} literature, by contrast, argues for maximizing policy entropy to support exploration until the optimal behavior is reliably discovered. Existing RLVR studies inherit one of these viewpoints in isolation. 
EMPO~\cite{zhang2025right} and one-shot RL~\cite{wang2025reinforcementlearningreasoninglarge} minimize predictive entropy to \revision{exploit the base model prior knowledge}, while CLIP-Cov~\cite{cui2025entropy_max_rlvr} prevents 
the collapse of the entropy, thus promoting exploration.


Both choices may break down under noisy supervision.  Let $\mathcal{L}_{\text{entropy}} $ be the token-level entropy loss defined in Eq.~\eqref{entropy_loss} and $\lambda$ be a positive constant. 
GRPO using $-\lambda \mathcal{L}_{\text{entropy}}$ as a regularization term in the total loss may drive the model to place overly high confidence on potentially incorrect labels. \revision{Furthermore, this minimization simultaneously} suppresses the response diversity that GRPO's group-wise normalization requires for \revision{informative} advantage estimation.
\revision{In contrast}, regularizing with 
$+\lambda \mathcal{L}_{\text{entropy}}$
alleviates over-confidence and preserves the alternative candidates \revision{necessary} for GRPO response diversity.
\revision{However, under consistent entropy maximization}, the policy struggles to converge because the probability mass \revision{is never encouraged to concentrate.}
\revision{Therefore, we argue that the direction of entropy optimization should not remain static. Rather, it should be  dynamically scheduled throughout the training process.
}\revision{
As illustrated in Figure~\ref{entropy_trend}, token-level entropy should be maximized early in training. This initial exploration resists overfitting to noisy labels and provides the response diversity necessary for informative advantage estimation. 
Later in training, the entropy should be minimized to transition the model from exploration to exploitation, allowing it to consolidate learned knowledge.}

\begin{table*}[t]
\centering
\small
\caption{Accuracy (\%) of Qwen2.5-VL-3B across different annotation noise levels on GUI grounding (ScreenSpot) and fine-grained classification (Pets37, 4-shot) tasks.}
\label{tab:complete_results}
\scalebox{0.85}{
\begin{tabular}{lcccccccccccccccc}
\toprule
& \multicolumn{8}{c}{\textbf{GUI Grounding}} & \multicolumn{8}{c}{\textbf{Fine-grained Classification}} \\
\cmidrule(lr){2-9} \cmidrule(lr){10-17}
\textbf{Method} & \textbf{Base} & \textbf{100\%} & \textbf{80\%} & \textbf{60\%} & \textbf{50\%} & \textbf{40\%} & \textbf{20\%} & \textbf{0\%} & \textbf{Base} & \textbf{100\%} & \textbf{80\%} & \textbf{60\%} & \textbf{50\%} & \textbf{40\%} & \textbf{20\%} & \textbf{0\%} \\
\midrule
Base Model & 70.6 & -- & -- & -- & -- & -- & -- &  & 59.2 & -- & -- & -- & -- & -- &-- & -- \\
GRPO &-- & 71.0 & 72.0 &75.8  &76.2  &79.8 &81.8 &82.2

&--& 54.7 &64.7 &67.3  &68.5  &68.8 &68.8 & \textbf{70.7} \\

GRPO w. Min. & -- & 73.2 & 75.2 & 77.4 & 77.4 & 77.6 & 79.0 & 79.0 

& -- & \textbf{59.3} & 64.6 & 66.9 & \textbf{68.6} & 68.7 &69.5 & 70.4\\

GRPO w. Max. & -- & 73.6 & 74.2 & 76.6 & 77.8 & \textbf{81.0} & \textbf{82.6}& 83.0 

& -- & 51.0 & 64.5 & 67.5 & 67.8 & 68.5 &68.9 &69.8 \\

GRPO w. Two. & -- & \textbf{75.8} & \textbf{77.0} & \textbf{79.4} & \textbf{80.2} & {80.6} & {82.4} & \textbf{83.6}

& -- & {54.3} & \textbf{65.5} & \textbf{67.5} & {68.4} & \textbf{69.0} &\textbf{69.7} &70.0 \\
\bottomrule
\end{tabular}
}
\end{table*}

Based on the above intuition, we propose a two-stage token-level entropy optimization framework for RLVR
training, thereby realizing the exploration-to-exploitation trajectory. Let $\mathcal{L}_{\text{GRPO}}$ denote the standard GRPO loss derived from the group-wise advantage formulation, and let $\lambda(\tau)$ be a scalar coefficient that varies with the training step $\tau$. The unified objective function is defined as: 



\begin{equation}
\label{eq:two_stage_loss}
\mathcal{L}_{\text{total}} = 
\mathcal{L}_{\text{GRPO}}
+ \lambda(\tau)\, \mathcal{L}_{\text{entropy}}.
\end{equation}


We define the schedule for \revision{$\lambda(\tau)$} as a simple piecewise function:
\begin{equation}
\label{eq:schedule}
\lambda(\tau)=
\begin{cases}
\;\;\lambda_{\max}, & \text{if } \tau \le \tau_{\text{switch}} \quad\text{(Stage 1: exploration)},\\[4pt]
-\lambda_{\min}, & \text{otherwise}\quad\text{(Stage 2: exploitation)},
\end{cases}
\end{equation}
with hyper-parameters \revision{$\lambda_{\max},\lambda_{\min}>0$}. 
During Stage 1, the positive coefficient instantiates an entropy maximization variant of GRPO, which encourages diverse sampling. \revision{The switching point is triggered at the $\tau_{\text{switch}}$ training step. We studied $\tau_{\text{switch}}$ in Section~\ref{ablation_study}.}
\revision{Subsequently, Stage 2 flips the coefficient $-\lambda_{\text{min}}$ to minimize entropy. This shifts the optimization objective, directing the model to produce confident outputs and consolidate the knowledge acquired during the exploration phase.} The adaptive scheduling ensures that the model benefits from both regimes. The pseudo code is given in Algorithm~\ref{alg:grpo}. 


%% file: sec/5_experiment.tex
\section{Experiments}

\subsection{Experimental Setup}
\textbf{Datasets and Training.}
We use GRPO~\cite{deepseekai2025deepseekv3technicalreport} to train base models. For Qwen-VL backbones, we adopt UI-R1 framework~\cite{luo2025guir1generalistr1style} for GUI grounding and Visual-RFT framework~\cite{liu2025visualrftvisualreinforcementfinetuning} for fine-grained classification. For Intern-VL backbones, we adopt verl-internvl framework \citep{wang2025internvl3, wang2024mpo}. For GUI grounding task, we randomly select 500 samples from ScreenSpot~\cite{cheng2024seeclickharnessingguigrounding} as a training set, with an equal distribution between mobile, web, and desktop. For fine-grained classification task, we utilize Pets37 \cite{parkhi2012cats} with 4-shot setting. 


\noindent\textbf{Evaluation.}
For evaluation of the GUI grounding task, we select 500 samples from ScreenSpot as a test set, which is different from the training samples but with the same platform distribution. For the fine-grained classification task, we use the official test split of the Pets37 dataset for evaluation. We adopt grounding and prediction accuracy as our evaluation metrics, which is calculated by matching the bounding box for the GUI grounding task and matching the label text for fine-grained classification. 
We compare five configurations: (1) Base pretrained model without RL (Base Model), (2) Standard GRPO (GRPO), (3) GRPO training \revision{with an additional entropy-minimization regularization term} (GRPO w. Min.), (4) GRPO with an additional entropy-maximization regularization term (GRPO w. Max.), and (5) Our proposed two-stage entropy-guided method (GRPO w. Two.).

\noindent\textbf{\revision{Noisy} Supervision Simulation.}
For the GUI grounding task, we simulate noisy labels by randomly generating a new bounding box in the image with the same size as the original ground truth bounding box, ensuring no overlap between them, and using it as the new noisy target. We reward the response if the grounding point is within the \revision{target} noisy bounding box.
For the fine-grained classification task, to create noisy annotations, we randomly replace the correct label with an incorrect one drawn from the remaining set of labels. We reward the response if the prediction matches the noisy label. Across both tasks, we generate datasets with noise levels $\{100\%, 80\%, 60\%, 50\%, 40\%, 20\%, 0\%\}$.

\begin{table*}[t]
\centering
\footnotesize
\caption{Effect of Various Base Models on the ScreenSpot Dataset. Accuracy (\%) of 4 backbones, Qwen2-VL-2B, Qwen2-VL-7B, Qwen2.5-VL-3B, InternVL-3.5-2B, trained with \revision{standard} GRPO versus the proposed two-stage entropy-guided method (i.e., \textbf{w. Two.}).}
\label{tab:screenspot_results_scaling}
\begin{tabular}{lcccccccccccc}
\toprule
 & \multicolumn{3}{c}{\textbf{Qwen2-VL-2B}} & \multicolumn{3}{c}{\textbf{Qwen2-VL-7B}} & \multicolumn{3}{c}{\textbf{Qwen2.5-VL-3B}} & \multicolumn{3}{c}{\textbf{InternVL-3.5-2B}} \\
\cmidrule(lr){2-4} \cmidrule(lr){5-7} \cmidrule(lr){8-10} \cmidrule(lr){11-13}
\textbf{Noise Level} & \textbf{Base} & \textbf{GRPO} & \textbf{w. Two.} & \textbf{Base} & \textbf{GRPO} & \textbf{w. Two.} & \textbf{Base} & \textbf{GRPO} & \textbf{w. Two.} & \textbf{Base} & \textbf{GRPO} & \textbf{w. Two.} \\
\midrule
\textbf{-}           & 11.2 & --    & --    & 37.2 & --    & --    & 70.6 & --    & --   & 48.6 & --    & --    \\
\textbf{100\%}       & -- & \textbf{17.0} & 14.4 & -- & \textbf{37.4} & 34.8 & -- & 71.0 & \textbf{75.8} & -- & 49.2 & \textbf{50.2} \\
\textbf{50\%}        & -- & \textbf{32.8} & 25.2 & -- & 61.2 & \textbf{69.8} & -- & 76.2 & \textbf{80.2} & -- & 56.8 & \textbf{59.2} \\
\textbf{20\%}        & -- & \textbf{50.0} & 44.4 & -- & 74.0 & \textbf{76.6} & -- & 81.8 & \textbf{82.4} & -- & 63.2 & \textbf{69.8} \\
\textbf{0\%}         & -- & 55.2 & \textbf{55.6} & -- & 75.4 & \textbf{78.0} & -- & 82.2 & \textbf{83.6} & -- & 66.8 & \textbf{69.8} \\
\bottomrule
\end{tabular}
\end{table*}

\begin{figure*}[t]
\centering
\includegraphics[width=1\linewidth]{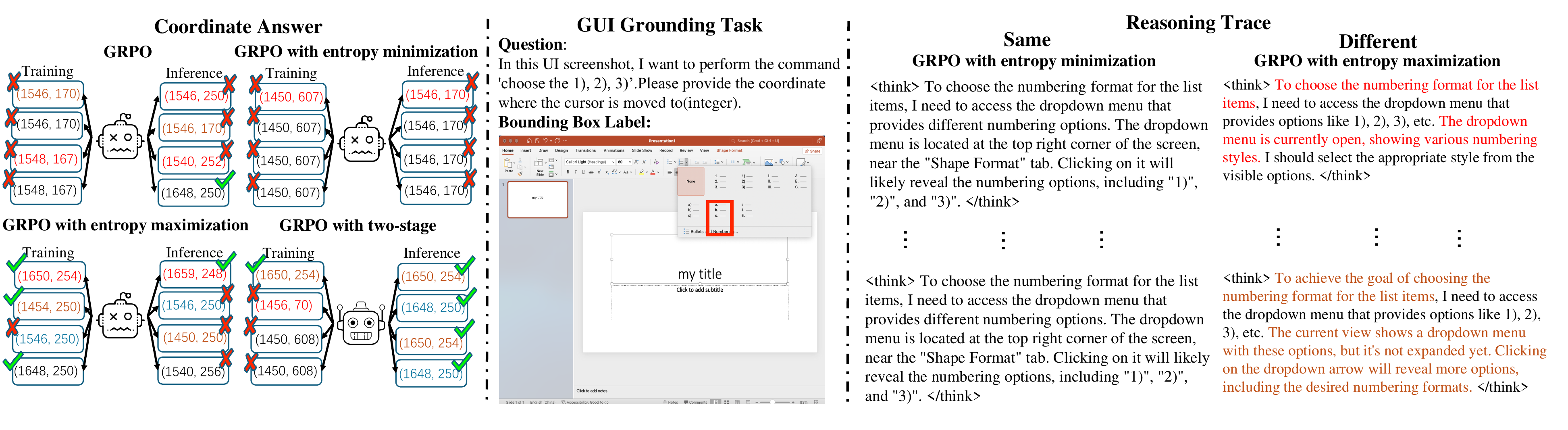}
\caption{Qualitative effect of entropy scheduling on the GUI grounding task. We visualise the reasoning trace (〈think〉…〈/think〉) and predicted coordinate produced by: GRPO, GRPO with entropy minimization, GRPO with entropy maximization, and GRPO with two-stage entropy\revision{-guided} optimization. The ground-truth bounding box is outlined in red on the image. 
}
\label{4_responses_4_setting}
\end{figure*}

\subsection{Main Results}

\textbf{Quantitative Analysis.}
Table~\ref{tab:complete_results} reports Qwen2.5-VL-3B's results on two different tasks. For GUI grounding, the proposed two-stage entropy\revision{-guided} optimization method maintains 80.2\% accuracy at 50\% noise, just 2\% below GRPO trained on clean labels, demonstrating remarkable noise tolerance. Our method consistently outperforms the standard GRPO baseline across all noise levels, with a particularly strong improvement of 4.8\% absolute gain at high noise (100\%). 
Absolute gains of 5.2\% to 13.0\% were achieved over the base Qwen2.5-VL-3B model under different noise conditions.
These results validate our core hypothesis that strategic entropy modulation enhances model performance under noisy data settings.
For fine-grained classification, the results share similar trends with GUI grounding. In particular, entropy minimization performs best at 100\% noise (59.3\%), while maximization excels at 0\% noise (69.8\%). Our method balances these regimes, delivering robust performance across noise levels. This confirms the task-agnostic benefits of our method.

Table~\ref{tab:screenspot_results_scaling} further reports results for three Qwen‐VL backbones and Internvl-3.5 on GUI grounding. Our two-stage method delivers gains across different model sizes, model families, and noise levels. 
Interestingly, we find that larger backbones \revision{benefit more} from the two–stage schedule, showing the potential scalability of our approach.
Qwen2-VL-7B records a significantly larger 8.6\% improvement at 50\% noise, while Qwen2-VL-2B has an opposite phenomenon. We also find our proposed approach has more significant gains on later models, e.g., Qwen2.5-VL-3B gains 4.8\% at 100\% noise and 4.2\% at 50\% noise over the GRPO baseline.
Beyond Qwen Model Family, we also train InternVL-3.5-2B on ScreenSpot, where the two-stage method consistently outperforms the standard GRPO baseline across all noise levels. This confirms that the benefits of phased entropy optimization are not limited to the Qwen model family. We include the full results of InternVL-3.5-2B in Appendix~\ref{additional_experiments}.

\noindent \textbf{Qualitative Analysis.}
Figure \ref{4_responses_4_setting} provides an illustrative comparison of how the three entropy regimes shape both the sampled reasoning traces and the final predictions. For GRPO with entropy minimization, the policy collapses almost immediately onto a single confident decoding path. All rollouts verbalize an almost identical chain of thought, so noisy rewards are propagated unchecked, and the model converges to the same incorrect coordinate at inference. In contrast, pure entropy maximization generates various reasoning paths that include at least one trajectory consistent with the true label, thus \revision{reducing the susceptibility to misleading reward signals}. However, the lack of consolidation leaves its accuracy short of the best. For our two-stage method, the reasoning traces remain diverse enough to resist noise but also coherent enough to pinpoint the correct GUI region.

\subsection{Discussions}

\textbf{Generalizable Findings.}
To further verify the general applicability of our method, we extend the study to the open-vocabulary object detection task (OVOD). Specifically, we randomly sampled 975 annotations from the COCO dataset \cite{lin2014microsoftcoco}, which includes 65 categories with 15 images per category. Similarly to GUI grounding task, we simulate label noise by generating bounding boxes that do not intersect with the original ground-truth boxes. Evaluation is performed on the remaining 15 categories that are unseen during training, using mean Average Precision (mAP) as the metric. We adopt the same GRPO method as in GUI grounding, with rewards computed based on box-overlap verification at an Intersection over Union (IoU) threshold of 0.5.
Table~\ref{tab:results_gsm_ovod} shows that the proposed two-stage entropy schedule significantly enhances the GRPO baseline across all noise conditions. Notably, at 50\% label noise, the two-stage approach improves the mAP of Qwen2-VL-2B by 3.53 from 15.94 (standard GRPO) to 19.47, matching the best score among all configurations.



\begin{table}[t]
\centering
\footnotesize
\caption{Performance comparison of Qwen2-VL-2B across different annotation noise levels on the OVOD task (mAP @ 0.5 IoU).}
\label{tab:results_gsm_ovod}
\scalebox{0.8}{
\begin{tabular}{lcccc}
\toprule
& \multicolumn{4}{c}{\textbf{OVOD}}  \\
\cmidrule(lr){2-5}
\textbf{Method} & \textbf{Base} & \textbf{100\%} & \textbf{50\%} & \textbf{0\%} \\
\midrule
Base Model  & 9.56  & --    & --             & --    \\
GRPO        & --    & 10.79 & 15.94          & 16.00 \\
GRPO w. Max.& --    & 14.60 & 19.47          & 17.20 \\
GRPO w. Min.& --    & \textbf{15.94} & 18.91 & \textbf{18.79} \\
GRPO w. Two.& --    & 15.54 & \textbf{19.47} & 18.44 \\
\bottomrule
\end{tabular}
}
\end{table}

\begin{table}[t]
\centering
\footnotesize
\caption{Performance comparison of Qwen2.5-VL-3B for the scaling effect of adding noisy training data to 500 clean GUI-grounding samples.}
\label{tab:ablation_add_noise}
\scalebox{0.9}{
\begin{tabular}{lcccccc}
\toprule
& \multicolumn{6}{c}{\textbf{ScreenSpot}} \\
\cmidrule(lr){2-7}
\textbf{Method} & \textbf{Base} & \textbf{+50} & \textbf{+100} &  \textbf{+150} & \textbf{+200} & \textbf{+250} \\
\midrule
Base Model  & 70.6    & --    & --    & --    & --    & --    \\
GRPO        & --    & 79.4  & 80.8  & 78.0  & 77.6  & 78.0  \\
GRPO w. Min.& --    & 79.8  & 79.4  & 78.6  & 79.6  & 79.0  \\
GRPO w. Max.& --    & \textbf{82.2}  & 81.4  & 82.0  & 80.4  & \textbf{80.4} \\
GRPO w. Two.& --    & {81.4} & \textbf{81.8} &\textbf{82.8} & \textbf{81.8} & 80.0 \\
\bottomrule
\end{tabular}
}
\end{table}

\noindent\textbf{Noisy Data Scaling.}
To further investigate the impact of noisy data on GRPO training, we fixed 500 \revision{correctly labeled} samples and \revision{incrementally} added 50 \revision{noisy samples} at a time to train models. 
As shown in Table \ref{tab:ablation_add_noise}, the standard GRPO baseline achieves \revision{its peak performance when 100 noisy samples are added to the 500 clean samples, after which accuracy begins to decline.} In contrast, our two-stage method maintains a highly robust 80.0\%-82.8\% accuracy across all noise scaling levels, demonstrating superior stability.
The noise effect is most pronounced for entropy maximization at +50 samples (82.2\%), but it degrades with additional noise. The consistent performance of our method confirms that the phased entropy optimization effectively \revision{exploits} noisy data benefits while mitigating its risks.

\noindent\textbf{Out-of-distribution Generalization.}
To assess the out-of-distribution (OOD) generalization ability, we evaluate on the ScreenSpot-Pro~\citep{li2025screenspotproguigroundingprofessional}, OS-World-G \cite{xie2025scalingcomputerusegroundinguser}, and MMBench-GUI L2~\citep{wang2025mmbench_gui} benchmarks, which differ significantly from the training distribution (ScreenSpot) in both visual complexity and domain coverage.
For ScreenSpot-Pro, we randomly sample 150 samples from each category (Development, Creative, CAD, Scientific, Office, OS) to ensure equal amount for each category. For MMBench-GUI L2, we randomly sample 500 samples while ensuring the data distribution is uniformed across six platforms (Windows, macOS, linux, iOS, Android, and Web). For OS-World-G, we use the whole dataset.

As shown in Table \ref{tab:ood}, the two-stage method achieves the best OOD performance (20.7\%) with 500 clean samples +150 mislabeled samples configuration (i.e., +150 configuration). This 2.7-5.4\% improvement over alternatives indicates that the two-stage entropy optimization method improves knowledge transfer. 
In particular, entropy maximization alone achieves competitive OOD performance at +50 samples (20.7\%) but degrades with additional noise, while our method maintains robust generalization. Across all noise levels on OS-World-G, our two-stage method (GRPO w. Two.) consistently delivers robust OOD performance, maintaining accuracy between 40.0\% and 42.4\%, outperforming standard GRPO (37.7\%–41.4\%) and single-stage variants (e.g., GRPO w. Max.: 38.7\%–42.6\%; GRPO w. Min.: 38.7\%–40.2\%).
We further evaluate our two-stage method for OOD generalization across ScreenSpot-Pro and MMBench-GUI L2, as shown in Table~\ref{tab:mmbench_gui_ood}. For instance, our two-stage method achieves the best overall OOD performance. It achieves 60.6\% accuracy on clean data and 57.4\% at 50\% noise, outperforming standard GRPO and single-stage entropy methods on MMBench-GUI L2. \revision{We include additional results on the MMBench-GUI L2 in Appendix~\ref{additional_experiments}.}

\begin{table}[t]
\centering
\caption{OOD evaluation accuracy (\%) of Qwen2.5-VL-3B trained on ScreenSpot, evaluated on ScreenSpot-Pro and OS-World-G across adding noisy training data configurations.}
\label{tab:ood}
\scalebox{0.58}{
\begin{tabular}{lcccccccccc}
\toprule
& \multicolumn{5}{c}{\textbf{ScreenSpot-Pro}} & \multicolumn{5}{c}{\textbf{OS-World-G}} \\
\cmidrule(lr){2-6} \cmidrule(lr){7-11}
\textbf{Method} & \textbf{+50} & \textbf{+100} & \textbf{+150} & \textbf{+200} & \textbf{+250} & \textbf{+50} & \textbf{+100} & \textbf{+150} & \textbf{+200} & \textbf{+250} \\
\midrule
GRPO        & 16.7          & 16.7          & 18.0          & 17.3          & \textbf{19.3} & 38.1 & 39.8 & 36.5  & 37.7 & \textbf{41.4} \\
GRPO w. Min.& 16.0          & 16.7          & 15.3          & 16.0          & 16.7          & 39.8 & 42.0  & 40.2 & 38.9 & 38.7 \\
GRPO w. Max.& \textbf{20.7} & 16.7          & 18.0          & 17.3          & 12.7          & \textbf{42.6} & 39.8 & \textbf{42.3} & 40.2 & 41.3 \\
GRPO w. Two.& 16.7          & \textbf{19.3} & \textbf{20.7} & \textbf{18.0} & 18.0          & 42.1 & \textbf{42.1} & 41.2 & \textbf{42.4} & 40.0 \\
\bottomrule
\end{tabular}
}
\end{table}

\begin{table}[t]
\centering
\caption{OOD evaluation accuracy (\%) of Qwen2.5-VL-3B trained on ScreenSpot, evaluated on ScreenSpot-Pro and MMBench-GUI L2 across different annotation noise levels.}
\label{tab:mmbench_gui_ood}
\scalebox{0.65}{
\begin{tabular}{lcccccccc}
\toprule
& \multicolumn{4}{c}{\textbf{ScreenSpot-Pro}} & \multicolumn{4}{c}{\textbf{MMBench-GUI L2}} \\
\cmidrule(lr){2-5} \cmidrule(lr){6-9}
\textbf{Method} & \textbf{Base} & \textbf{0\%} & \textbf{50\%} & \textbf{100\%} & \textbf{Base} & \textbf{0\%} & \textbf{50\%} & \textbf{100\%} \\
\midrule
Base Model    & 6.4   & --             & --             & --             & 45.0 & --             & --             & --            \\
GRPO          & --   & 16.7           & 13.3           & 8.7           & --   & 57.0           & 53.8           & 46.4           \\
GRPO w.\ Min. & --   & 18.7           & 11.3           & \textbf{8.7}  & --   & 58.0           & 55.6           & \textbf{51.0}  \\
GRPO w.\ Max. & --   & 16.7           & 14.0           & 7.3           & --   & 58.2           & 54.6           & 48.8           \\
GRPO w.\ Two. & --   & \textbf{21.3}  & \textbf{18.0}  & 8.0           & --   & \textbf{60.6}  & \textbf{57.4}  & 49.0           \\
\bottomrule
\end{tabular}
}
\end{table}


\noindent\textbf{GRPO tolerance to Data Noise.}
Fig.~\ref{cfa} and Table~\ref{tab:complete_results} reveal that standard GRPO already exhibits \revision{moderate} robustness to noisy labels. With 50\% noisy GUI‐grounding labels, Qwen2.5-VL-3B trained with standard GRPO attains 76.2\% accuracy, only 6\% below the clean-data ceiling. \revision{We hypothesize that this robustness arises partly from GRPO’s group-wise advantage normalization. When evaluating a mislabeled sample, if the model's prior ability leads all $K$ rollouts to consistently predict the actual correct answer}, every response in the group receives the same zero reward. Consequently, the normalized advantages become zero, which yields no learning signal for that group. 
This self-gating effect establishes a robust baseline on top of which entropy scheduling can operate.
\revision{To test whether the noise tolerance observed in GRPO is an inherent algorithmic property rather than an artifact of data formatting or preprocessing, we conduct additional ablation studies in Appendix~\ref{additional_experiments}. 
Results show GRPO remains robust to noisy labels across preprocessing variations.}

\begin{figure}[t]
\centering
\includegraphics[width=1\linewidth]{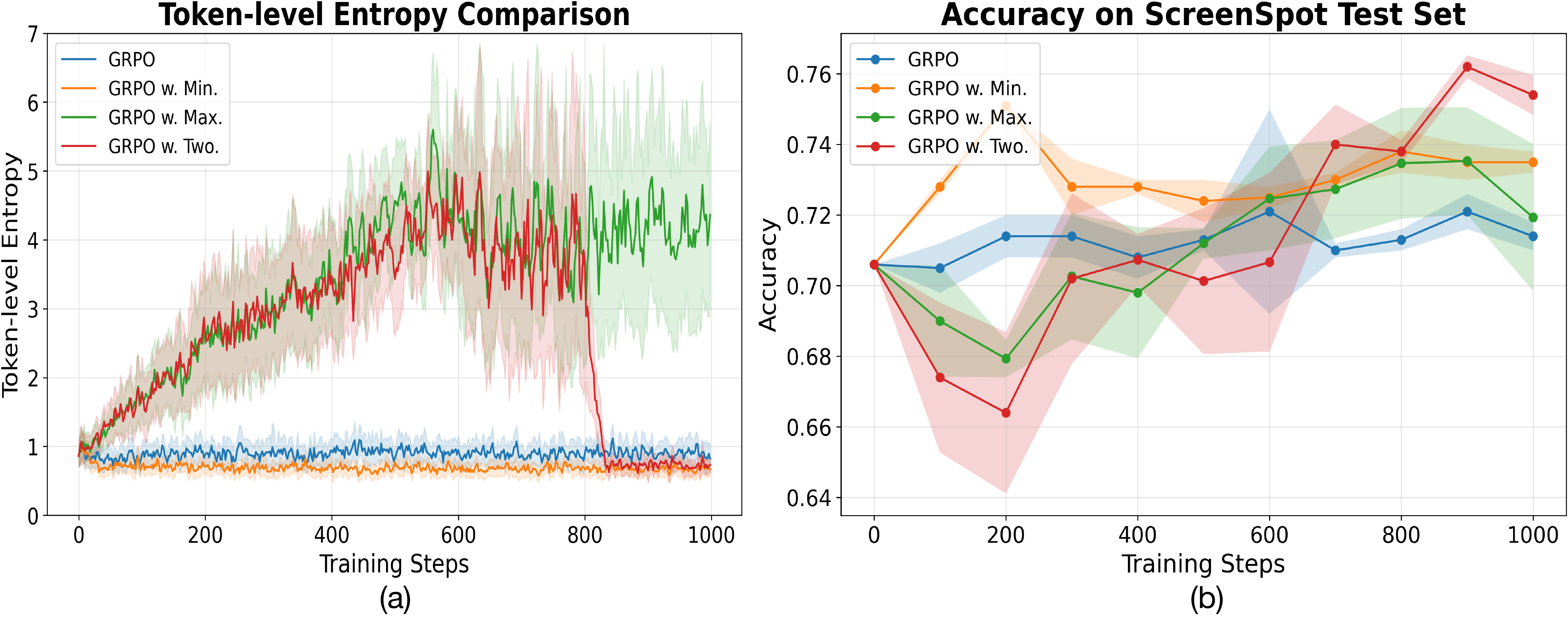}
\caption{(a) Comparison of token-level entropy dynamics during training with 100\% noise; (b) Comparison of ScreenSpot test set accuracy at each training step under 100\% noise level. We compare 4 strategies: standard GRPO, GRPO with entropy maximization, GRPO with entropy minimization, and GRPO with two-stage entropy\revision{-guided} optimization. }
\label{entropy_trend}
\end{figure}

\subsection{Ablation Study}
\label{ablation_study}



\noindent\textbf{Training Dynamics of Entropy.}
For our proposed two-stage entropy-guided optimization method, Figure \ref{entropy_trend} illustrates the training \revision{dynamics} of token-level entropy. During Phase 1 (steps 0-800), the entropy increases steadily to \revision{$\sim$}400\% of the initial value, confirming effective exploration. The transition to Phase 2 (steps 800-1000) triggers \revision{a rapid reduction} in entropy, stabilizing at $\sim$20\% \ of the peak value after 900 steps. 
These dynamics validate our core design: extended exploration prevents premature convergence, while subsequent exploitation consolidates knowledge into confident predictions. This smooth phase transition is crucial for maintaining stability under noisy supervision. 

\begin{table}[t]
\centering
\footnotesize
\caption{Influence of the exploration-to-exploitation switch point for Qwen2.5-VL-3B on the GUI grounding task.}
\label{tab:two-stage transition point}
\scalebox{1}{
\begin{tabular}{lcccc}
\toprule
                     &               \multicolumn{4}{c}{\textbf{Transition Point}}\\
\cmidrule(lr){2-5}
\textbf{Noise Level}      & \textbf{Step 500}& \textbf{Step 700}& \textbf{Step 800}& \textbf{Step 900}\\
\midrule
100\%& 73.6& 75.0& \textbf{75.8}& 73.6\\
50\%& 79.6& 79.8& \textbf{80.2}& 79.0\\
0\%& 80.4& 81.8& \textbf{83.6}& 82.0\\
\bottomrule

\end{tabular}}
\end{table}

\noindent\textbf{Switching Points Analysis.}
We examine the effect of switching point on the GUI grounding task (ScreenSpot as the training set), by varying $ \tau_{\text{switch}} \in \{500, 700, 800, 900\}$. As shown in Table~\ref{tab:two-stage transition point}, the performance is generally robust to the fixed switching point. Switching at step 800, corresponding to 80\% of the total training steps, achieves the best balance between sufficient exploration and late-stage consolidation on GUI grounding.

\begin{table}[t]
\footnotesize
\centering
\caption{Performance Comparison Across Two-stage Methods for Qwen2.5-VL-3B on the GUI grounding task. LT. refers to training samples with correct labels. LF. refers to training samples with incorrect labels.  LT. Max. LF. Min. refers to maximizing entropy on the correctly-labeled subset and minimize it on the noisy subset.}
\label{tab:two-stage method}
\scalebox{1}{
\begin{tabular}{lccc}
\toprule
                     &               \multicolumn{3}{c}{\textbf{Noise Level}}\\
\cmidrule(lr){2-4}
\textbf{Methods}& \textbf{100\%}& \textbf{50\%}& \textbf{0\%}\\
\midrule

\textbf{LT. Max. LF. Min.}& 73.2& 76.8& 83.0\\
 \textbf{LF. Max. LT. Min.}& 73.6& 78.0& 79.0\\
\textbf{Min. then Max.}& 70.2& 76.8& 79.8\\
 \textbf{Max. then Min.}& \textbf{75.8}
& \textbf{80.2}
& \textbf{83.6}\\
\bottomrule
\end{tabular}}
\end{table}

\noindent\revision{\textbf{Stage-wise entropy scheduling (“Max. then Min.”) outperforms subset-wise entropy assignment.}  }
Table~\ref{tab:two-stage method} compares four possible ways of combining entropy maximization and minimization under 100\%, 50\% and 0\% noise level (ScreenSpot as the training set). 
Across all noise levels, “Max. then Min.” outperforms “Min. then Max.” by 5.6\% at 100\% noise level, 3.4\% at 50\% noise level and 3.8\% on clean data. \revision{Beginning with entropy minimization prematurely encourages the policy to converge. This causes the model to overfit to the initial noisy reward signals, reducing the response diversity required for GRPO to discover better trajectories later in training.} Conversely, starting with entropy maximization supports \revision{the exploration and} the diversity needed for effective group-wise advantage estimation in GRPO. \revision{The subsequent minimization phase then consolidates the high reward behaviors discovered during the exploration phase into a confident policy.}

When maximizing entropy is restricted to the noisy subset only (i.e.,“LF. Max. LT. Min.”), its performance is better than “Min. then Max.” but still inferior to “Max. then Min.” schedule. \revision{Applying entropy maximization exclusively to the noisy subset restricts the model's overall ability to explore. Even for correctly labeled data, initial exploration is beneficial for discovering potentially better reasoning paths before committing to a final policy. Furthermore, in practical settings, the clean or noisy status of a label is unknown, making a unified schedule much more applicable.} Symmetrically, “LT. Max. LF. Min.” works well with 50\% and 0\% noise levels, because half or all data are reliable, but it suffers under 100\% noise level when there are no clean labels to guide the exploitation.

%% file: sec/6_discussion.tex
\section{Conclusion}


We explore the effectiveness of RLVR under noisy supervision for multimodal tasks. To augment RLVR methods like GRPO, we propose a Two-Stage Entropy-Guided GRPO that first maximizes and then minimizes the token-level entropy during training. This strategy encourages early exploration and later exploitation, leading to improved robustness against label noise. Through extensive experiments with Qwen \revision{and InternVL} models and on different tasks, we demonstrate that our method maintains high performance even under substantial annotation noise. In particular, the two-stage method contributes to more stable convergence and better generalization. Our findings highlight the potential of entropy-aware policy optimization as a powerful tool for learning from imperfect data in multimodal scenarios.

%% file: sec/X_suppl.tex
\clearpage
\setcounter{page}{1}
\maketitlesupplementary

\counterwithin{table}{section}

\section{Limitations.}
A limitation of the Two-Stage Entropy-Guided GRPO approach is that it works best when the base model has a reasonable prior ability on the target task. If the zero-shot ability of the base model in the target task is weak, early maximization of entropy can amplify incorrect modes before the model samples a correct trajectory. \revision{This likely explains the weaker gains for Qwen2-VL-2B in Table~\ref{tab:screenspot_results_scaling} and the limited benefit under fully noisy supervision on fine-grained classification in Table~\ref{tab:complete_results}.}

\section{Implementation Details}
\textbf{Training Details.}
We provide a brief summary of the training settings in Table \ref{tab:hyperparams}. For both the GUI grounding and fine-grained classification tasks, the base model is trained using 8 NVIDIA L20 GPUs, requiring approximately 8 hours and 1 hour, respectively. OVOD tasks share the same setting as fined-grained classification tasks. Code website: \url{https://github.com/xudonglai0426/RLVR-from-Exploration-to-Exploitation}.

\begin{table}[h]
\centering
\caption{Hyperparameter settings used in the experiments.}
\label{tab:hyperparams}
\scalebox{0.85}{
\begin{tabular}{@{}ccc@{}}\toprule

\textbf{Hyperparameter} & \textbf{GUI Ground.} & \textbf{Fine. Class.} \\\midrule

Learning rate (lr)          & 9.98 $\times$ 10$^{-7}$ to 0 & 9.98 $\times$ 10$^{-7}$ to 0 \\
Max pixels                  & 12,845,056 & 401,408\\
Number of generations       & 8 & 8\\
Number of training epochs   & 4 & 24\\
Max prompt length           & 1024 & 1024\\
Per-device train batch size & 1 & 1\\
Gradient accumulation steps & 2 & 2\\
Entropy Coef. & $1\times 10^{-2}$ & $1\times 10^{-2}$ \\ 
 KL Coef.& $4\times 10^{-2}$&0\\ \bottomrule 
\end{tabular}
}
\end{table}
\noindent\textbf{Evaluation Details.}
For the MMBench-GUI L2 benchmark, we randomly sample 500 samples for the training set and 500 samples for the test set. Both sets share the same data composition, with an equal distribution across the six platforms (Windows, macOS, Linux, iOS, Android, and Web) and the two instruction types (basic and advanced). For OS-World-G benchmark, we use the whole dataset with refined instruction for evaluation. 

\section{Entropy Optimization Schedule}

\textbf{Why Training Starts with Entropy Maximization.}
Our two-stage schedule begins with token-level entropy maximization because diversity is the currency that GRPO relies on to compute meaningful advantage signals. Maximization enlarges the variance of responses within each group, sharpening the relative ranking and, consequently, the gradient. At the same time, it regularizes the policy against premature converge to spurious labels. When the correct supervision is missing or wrong, a more diverse distribution prevents the policy from overfitting to the noisy target. Empirically, this exploration phase already yields a non-trivial improvement over either entropy minimization or the plain GRPO baseline (e.g. 77.8\% vs. 76.2\% at 50\% noise on ScreenSpot).

\noindent\textbf{Why Training ends with Entropy Minimization.}
Exploration alone is insufficient. Once the policy has discovered high-reward regions, it must consolidate. After token entropy plateaus, the sign of the entropy coefficient is flipped. Minimizing entropy concentrates probability mass on the best trajectory identified earlier, reduces variance at inference time and sharpens predictions. The switch consistently achieves improvements across all noise levels, confirming that exploitation effectively complements exploration.

\begin{algorithm}[t]
\caption{Two-Stage Entropy-\revision{Guided} GRPO}
\label{alg:grpo}
\small
\begin{algorithmic}[1]
\State \textbf{Require:} switch step $\tau_{\text{switch}}$, coefficients $\lambda_{\max}$, $\lambda_{\min}$, total training steps $E$, model $\pi_\theta$ with parameters $\theta$.
\For{$\tau = 1$ \textbf{to} $E$}
    \State Sample $K$ responses $\{y_1, \ldots, y_K\}$ from $\pi_\theta(\cdot|x)$
    \State Compute rewards \revision{$r_i=\mathcal{R}(y_i, y^*)$} for each response
    \State Compute normalized advantages:
  \revision{  \[
    A_i = \frac{r_i - \text{mean}(r( y_{1:K}))}{\text{std}(r(y_{1:K}))}
    \]}
    \If{$\tau \le \tau_{\text{switch}}$}
        \State $\lambda(\tau) \leftarrow +\lambda_{\max}$ 
    \Else
        \State $\lambda(\tau) \leftarrow -\lambda_{\min}$ 
    \EndIf
    \State Compute standard GRPO loss: 
    $
    \mathcal{L}_{\text{GRPO}}
    $ (see Eq.~\eqref{grpo_loss})
    \State Compute entropy regularization term:
    $
    \mathcal{L}_{\text{entropy}}
    $ (see Eq.~\eqref{entropy_loss})
    \State Compute total loss:
    $
    \mathcal{L}_{\text{total}}
    $ (see Eq.~\eqref{eq:two_stage_loss})
    \State Update $\theta$ with AdamW on $\nabla_\theta \mathcal{L}_{\text{total}}$
\EndFor
\State \textbf{return} trained model $\pi_\theta$
\end{algorithmic}
\end{algorithm}

\begin{table*}[t]
\centering
\footnotesize
\caption{\revision{Robustness Analysis of GRPO on ScreenSpot at 100\% noise level.}}
\label{tab:screenspot_variants_resizing}
\scalebox{0.9}{
\begin{tabular}{lcccccc}
\toprule
\textbf{Model} & \textbf{GRPO} & \textbf{GRPO with Two.} & \textbf{GRPO w. Abs. Coord.} & \textbf{GRPO w. Two. and Abs. Coord.} & \textbf{GRPO w. Resize} & \textbf{GRPO w. Two. w. Resize}\\
\midrule
Qwen2-VL-2B    & 12.4 & 13.8& 14.5 & 16.0& 13.4 & 16.6\\
Qwen2.5-VL-3B  & 69.8 & 73.8& 69.8& 73.8& 70.6 & 74.2\\
InternVL3.5-2B & 49.2 & 50.2& 49.2& 50.2& 46.2 & 49.8\\
\bottomrule
\end{tabular}
}
\end{table*}

\section{Additional Experiments}
\label{additional_experiments}


\begin{table}[t]
\centering
\footnotesize
\caption{\revision{Accuracy (\%) of InternVL-3.5-2B across annotation noise levels
on the GUI grounding (ScreenSpot) task.}}
\label{tab:internvl_screenspot}
\scalebox{0.9}{
\begin{tabular}{lcccccc}
\toprule
\textbf{Method} & \textbf{Base} & \textbf{100\%} & \textbf{80\%} & \textbf{50\%} & \textbf{20\%} & \textbf{0\%} \\
\midrule
Base Model    & 48.6& - & - & - & - & - \\
GRPO          & - & 49.2          & 49.6          & 56.8          & 63.2          & 66.8          \\
GRPO w.\ Min. & - & 48.0          & 51.0          & \textbf{59.8} & 65.6          & 69.2          \\
GRPO w.\ Max. & - & 49.8          & 51.6          & 57.6          & 66.0          & 65.2          \\
GRPO w.\ Two. & - & \textbf{50.2} & \textbf{53.0} & 59.2          & \textbf{69.8} & \textbf{69.8} \\
\bottomrule
\end{tabular}
}
\end{table}

\begin{table}[t]
\centering
\footnotesize
\caption{In-domain training Accuracy (\%) on MMBench-GUI L2 of Qwen2.5-VL-3B. The model is trained on MMBench-GUI L2 under \{100\%, 50\%, 0\%\} annotation noise.}
\label{tab:mmbench_gui_main_in_domain}
\begin{tabular}{lcccc}
\toprule
\textbf{Method} & \textbf{Base} & \textbf{100\%} & \textbf{50\%} & \textbf{0\%} \\
\midrule
Base Model      & 45.0 & - & - & - \\
GRPO            & -  & 47.0          & 53.6          & 55.0          \\
GRPO w.\ Min.   & -  & 51.0          & 54.6          & 57.6 \\
GRPO w.\ Max.   & -  & 49.6 & 56.0 & 58.0          \\
GRPO w.\ Two.   & -  & 49.4          & 53.8          & 55.0        \\
\bottomrule
\end{tabular}
\end{table}

\begin{table}[t]
\centering
\footnotesize
\caption{In-domain training Accuracy (\%) on GSM8K of Qwen2.5-3B. The model is trained on GSM8K under \{100\%, 50\%, 0\%\} annotation noise.}
\label{tab:gsm8k_in_domain}
\begin{tabular}{lcccc}
\toprule
\textbf{Method} & \textbf{Base} & \textbf{100\%} & \textbf{50\%} & \textbf{0\%} \\
\midrule
Base Model      & 77.2& - & - & - \\
GRPO            & -  & 80.4& 78.6& 81.4\\
GRPO w.\ Min.   & -  & 80.4& 81.4& 83.2\\
GRPO w.\ Max.   & -  & 81& 81.6& 80.6\\
GRPO w.\ Two.   & -  & 80.4& 80.6& 79.6\\
\bottomrule
\end{tabular}
\end{table}

\paragraph{\revision{Robust Analysis of GRPO.}}
As observed in Fig.~\ref{cfa} and Table~\ref{tab:complete_results}, the standard GRPO already exhibits moderate robustness to noisy labels. To address potential concerns that this noise tolerance might be an artifact of specific data preprocessing choices, we conducted an ablation study on coordinate formatting and image scaling with 4 rollouts during training. Specifically, we evaluated the GRPO baseline under 100\% noise using absolute coordinates (GRPO w. Abs. Coord.) instead of relative ones, and with dynamic image resizing enabled (GRPO w. Resize). As shown in Table~\ref{tab:screenspot_variants_resizing}, performance remains stable across these preprocessing variations. This confirms that the noise tolerance is an inherent algorithmic property of GRPO. Specifically, the self-gating effect where uniform incorrect predictions within a group yield zero normalized advantage mitigates harmful gradient updates.

\paragraph{\revision{Evaluation on MMBench-GUI L2.}}

To resolve potential concerns about training data contamination, we conducted additional experiments using the MMBench-GUI L2 dataset. Since MMBench-GUI L2 was published after the knowledge cutoff of the Qwen2.5-VL-3B base model, it serves as an ideal benchmark for data contamination evaluation. We evaluate our approach under in-domain training settings.
For in-domain training on the MMBench-GUI L2, we train and evaluate the model directly on the MMBench-GUI L2 dataset under different annotation noise levels. We set the transition step as 400 and evaluate at training step 500. As shown in Table~\ref{tab:mmbench_gui_main_in_domain}, our two-stage method achieves 49.4\% accuracy, outperforming the base model (45.0\%) and standard GRPO (47.0\%).

\paragraph{\revision{Experiments on Text-based Tasks.}}

To further investigate our two-stage method, we conducted additional experiments using the GSM8K \cite{cobbe2021gsm8k} dataset with Qwen2.5-3B \cite{qwen2.5} in Table~\ref{tab:gsm8k_in_domain}. We randomly select 500 samples from the training set and 500 samples from the test set of the full dataset for training and evaluation, respectively.